\documentclass{article}

\PassOptionsToPackage{numbers,compress}{natbib}
\usepackage[preprint]{neurips_2026}

\usepackage[utf8]{inputenc}
\usepackage[T1]{fontenc}
\usepackage{hyperref}
\usepackage{url}
\usepackage{booktabs}
\usepackage{array}
\usepackage{amsmath}
\usepackage{amsfonts}
\usepackage{multirow}
\usepackage{nicefrac}
\usepackage{microtype}
\usepackage{xcolor}
\usepackage{colortbl}
\usepackage{enumitem}
\usepackage{inconsolata}
\usepackage{listings}
\usepackage{fancyvrb}
\usepackage{tikz}
\usepackage{fontawesome5}
\usetikzlibrary{arrows.meta,positioning,calc,shadows}

\definecolor{keywordcolor}{rgb}{0.7,0.1,0.1}
\definecolor{tacticcolor}{rgb}{0.0,0.1,0.6}
\definecolor{commentcolor}{rgb}{0.4,0.4,0.4}
\definecolor{symbolcolor}{rgb}{0.0,0.1,0.6}
\definecolor{sortcolor}{rgb}{0.1,0.5,0.1}
\definecolor{attributecolor}{rgb}{0.7,0.1,0.1}

\lstdefinelanguage{lean} {
mathescape=false,
texcl=false,
morekeywords=[1]{
import, prelude, protected, private, noncomputable, definition, meta, renaming,
hiding, parameter, parameters, begin, constant, constants,
lemma, variable, variables, theory,
print, theorem, example,
open, as, export, override, axiom, axioms, inductive, with,
structure, record, universe, universes,
alias, help, precedence, reserve, declare_trace, add_key_equivalence,
match, infix, infixl, infixr, notation, postfix, prefix, instance,
eval, reduce, check, end, this,
using, using_well_founded, namespace, section,
attribute, local, set_option, extends, include, omit, class,
raw, replacing,
calc, have, show, suffices, by, in, at, let, forall, Pi, fun,
exists, if, dif, then, else, assume, obtain, from, register_simp_ext, unless, break, continue,
mutual, do, def, run_cmd, const,
partial, mut, where, macro, syntax, deriving,
return, try, catch, for, macro_rules, declare_syntax_cat, abbrev, sorry},
morekeywords=[2]{Sort, Type, Prop},
morekeywords=[3]{
assumption,
apply, intro, intros, allGoals, 
generalize, clear, revert, done, exact,
refine, repeat, cases, rewrite, rw,
simp, simp_all, contradiction,
constructor, injection,
induction, 
prove_with, linarith, nlinarith, norm_num, contrapose, rfl, by_contra, push_neg, gcongr, aesop, field_simp, rfl, norm_cast, ring, ring_nf, positivity, omega, subst, rename_i, rcases, convert, ext, tauto, native_decide, use, by_cases,
to_theorem, operatorcount, quickcheck,
},
literate=
{α}{{\ensuremath{\mathrm{\alpha}}}}1
{β}{{\ensuremath{\mathrm{\beta}}}}1
{γ}{{\ensuremath{\mathrm{\gamma}}}}1
{δ}{{\ensuremath{\mathrm{\delta}}}}1
{ε}{{\ensuremath{\mathrm{\varepsilon}}}}1
{ζ}{{\ensuremath{\mathrm{\zeta}}}}1
{η}{{\ensuremath{\mathrm{\eta}}}}1
{θ}{{\ensuremath{\mathrm{\theta}}}}1
{ι}{{\ensuremath{\mathrm{\iota}}}}1
{κ}{{\ensuremath{\mathrm{\kappa}}}}1
{μ}{{\ensuremath{\mathrm{\mu}}}}1
{ν}{{\ensuremath{\mathrm{\nu}}}}1
{ξ}{{\ensuremath{\mathrm{\xi}}}}1
{π}{{\ensuremath{\mathrm{\mathnormal{\pi}}}}}1
{ρ}{{\ensuremath{\mathrm{\rho}}}}1
{σ}{{\ensuremath{\mathrm{\sigma}}}}1
{τ}{{\ensuremath{\mathrm{\tau}}}}1
{φ}{{\ensuremath{\mathrm{\varphi}}}}1
{χ}{{\ensuremath{\mathrm{\chi}}}}1
{ψ}{{\ensuremath{\mathrm{\psi}}}}1
{ω}{{\ensuremath{\mathrm{\omega}}}}1
{Γ}{{\ensuremath{\mathrm{\Gamma}}}}1
{Δ}{{\ensuremath{\mathrm{\Delta}}}}1
{Θ}{{\ensuremath{\mathrm{\Theta}}}}1
{Λ}{{\ensuremath{\mathrm{\Lambda}}}}1
{Σ}{{\ensuremath{\mathrm{\Sigma}}}}1
{Φ}{{\ensuremath{\mathrm{\Phi}}}}1
{Ξ}{{\ensuremath{\mathrm{\Xi}}}}1
{Ψ}{{\ensuremath{\mathrm{\Psi}}}}1
{Ω}{{\ensuremath{\mathrm{\Omega}}}}1
{ℵ}{{\ensuremath{\aleph}}}1
{≤}{{\ensuremath{\leq}}}1
{≥}{{\ensuremath{\geq}}}1
{≠}{{\ensuremath{\neq}}}1
{≈}{{\ensuremath{\approx}}}1
{≡}{{\ensuremath{\equiv}}}1
{≃}{{\ensuremath{\simeq}}}1
{≤}{{\ensuremath{\leq}}}1
{≥}{{\ensuremath{\geq}}}1
{∂}{{\ensuremath{\partial}}}1
{∆}{{\ensuremath{\triangle}}}1 
{∫}{{\ensuremath{\int}}}1
{∑}{{\ensuremath{\mathrm{\Sigma}}}}1
{Π}{{\ensuremath{\mathrm{\Pi}}}}1
{⊥}{{\ensuremath{\perp}}}1
{∞}{{\ensuremath{\infty}}}1
{∂}{{\ensuremath{\partial}}}1
{∓}{{\ensuremath{\mp}}}1
{±}{{\ensuremath{\pm}}}1
{×}{{\ensuremath{\times}}}1
{⊕}{{\ensuremath{\oplus}}}1
{⊗}{{\ensuremath{\otimes}}}1
{⊞}{{\ensuremath{\boxplus}}}1
{∇}{{\ensuremath{\nabla}}}1
{√}{{\ensuremath{\sqrt}}}1
{⬝}{{\ensuremath{\cdot}}}1
{•}{{\ensuremath{\cdot}}}1
{∘}{{\ensuremath{\circ}}}1
{⁻}{{\ensuremath{^{-}}}}1
{▸}{{\ensuremath{\blacktriangleright}}}1
{∧}{{\ensuremath{\wedge}}}1
{∨}{{\ensuremath{\vee}}}1
{¬}{{\ensuremath{\neg}}}1
{⊢}{{\ensuremath{\vdash}}}1
{⟨}{{\ensuremath{\langle}}}1
{⟩}{{\ensuremath{\rangle}}}1
{↦}{{\ensuremath{\mapsto}}}1
{←}{{\ensuremath{\leftarrow}}}1
{<-}{{\ensuremath{\leftarrow}}}1
{→}{{\ensuremath{\rightarrow}}}1
{↔}{{\ensuremath{\leftrightarrow}}}1
{⇒}{{\ensuremath{\Rightarrow}}}1
{⟹}{{\ensuremath{\Longrightarrow}}}1
{⇐}{{\ensuremath{\Leftarrow}}}1
{⟸}{{\ensuremath{\Longleftarrow}}}1
{∩}{{\ensuremath{\cap}}}1
{∪}{{\ensuremath{\cup}}}1
{⊂}{{\ensuremath{\subseteq}}}1
{⊆}{{\ensuremath{\subseteq}}}1
{⊄}{{\ensuremath{\nsubseteq}}}1
{⊈}{{\ensuremath{\nsubseteq}}}1
{⊃}{{\ensuremath{\supseteq}}}1
{⊇}{{\ensuremath{\supseteq}}}1
{⊅}{{\ensuremath{\nsupseteq}}}1
{⊉}{{\ensuremath{\nsupseteq}}}1
{∈}{{\ensuremath{\in}}}1
{∉}{{\ensuremath{\notin}}}1
{∋}{{\ensuremath{\ni}}}1
{∌}{{\ensuremath{\notni}}}1
{∅}{{\ensuremath{\emptyset}}}1
{∖}{{\ensuremath{\setminus}}}1
{†}{{\ensuremath{\dag}}}1
{ℕ}{{\ensuremath{\mathbb{N}}}}1
{ℤ}{{\ensuremath{\mathbb{Z}}}}1
{ℝ}{{\ensuremath{\mathbb{R}}}}1
{ℚ}{{\ensuremath{\mathbb{Q}}}}1
{ℂ}{{\ensuremath{\mathbb{C}}}}1
{⌞}{{\ensuremath{\llcorner}}}1
{⌟}{{\ensuremath{\lrcorner}}}1
{⦃}{{\ensuremath{\{\!|}}}1
{⦄}{{\ensuremath{|\!\}}}}1
{‖}{{\ensuremath{\|}}}1
{₁}{{\ensuremath{_1}}}1
{₂}{{\ensuremath{_2}}}1
{₃}{{\ensuremath{_3}}}1
{₄}{{\ensuremath{_4}}}1
{₅}{{\ensuremath{_5}}}1
{₆}{{\ensuremath{_6}}}1
{₇}{{\ensuremath{_7}}}1
{₈}{{\ensuremath{_8}}}1
{₉}{{\ensuremath{_9}}}1
{₀}{{\ensuremath{_0}}}1
{ᵢ}{{\ensuremath{_i}}}1
{ⱼ}{{\ensuremath{_j}}}1
{ₐ}{{\ensuremath{_a}}}1
{¹}{{\ensuremath{^1}}}1
{ₙ}{{\ensuremath{_n}}}1
{ₘ}{{\ensuremath{_m}}}1
{ₚ}{{\ensuremath{_p}}}1
{↑}{{\ensuremath{\uparrow}}}1
{↓}{{\ensuremath{\downarrow}}}1
{...}{{\ensuremath{\ldots}}}1
{·}{{\ensuremath{\cdot}}}1
{▸}{{\ensuremath{\triangleright}}}1
{∣}{{\ensuremath{\mid}}}1
{¹}{{\ensuremath{^1}}}1
{²}{{\ensuremath{^2}}}1
{³}{{\ensuremath{^3}}}1
{⁴}{{\ensuremath{^4}}}1
{⁵}{{\ensuremath{^5}}}1
{⁶}{{\ensuremath{^6}}}1
{⁷}{{\ensuremath{^7}}}1
{⁸}{{\ensuremath{^8}}}1
{⁹}{{\ensuremath{^9}}}1
{⁰}{{\ensuremath{^0}}}1
{Σ}{{\color{symbolcolor}\ensuremath{\Sigma}}}1
{Π}{{\color{symbolcolor}\ensuremath{\Pi}}}1
{∀}{{\color{symbolcolor}\ensuremath{\forall}}}1
{∃}{{\color{symbolcolor}\ensuremath{\exists}}}1
{λ}{{\color{symbolcolor}\ensuremath{\mathrm{\lambda}}}}1
{\$}{{\color{symbolcolor}\$}}1
{:=}{{\color{symbolcolor}:=}}1
{=}{{\color{symbolcolor}=}}1
{<|>}{{\color{symbolcolor}<|>}}1
{<\$>}{{\color{symbolcolor}<\$>}}1
{+}{{\color{symbolcolor}+}}1
{*}{{\color{symbolcolor}*}}1,
morecomment=[s][\color{commentcolor}]{/-}{-/},
morecomment=[l][\itshape \color{commentcolor}]{--},
showstringspaces=false,
keepspaces=true,
morestring=[b]",
morestring=[d],
tabsize=3,
extendedchars=false,
sensitive=true,
breaklines=true,
breakatwhitespace=true,
basicstyle=\ttfamily\small,
captionpos=b,
columns=[l]fullflexible,
identifierstyle={\ttfamily\color{black}},
keywordstyle=[1]{\ttfamily\color{keywordcolor}},
keywordstyle=[2]{\ttfamily\color{sortcolor}},
keywordstyle=[3]{\ttfamily\color{tacticcolor}},
keywordstyle=[4]{\ttfamily\color{attributecolor}},
stringstyle=\ttfamily,
commentstyle={\ttfamily\footnotesize },
}

\newcommand{\lean}[1]{\lstinline[language=lean,basicstyle=\ttfamily]!#1!}

\usepackage{tcolorbox}
\tcbuselibrary{listings,skins,breakable}
\definecolor{codebg}{HTML}{F5F5F5}
\definecolor{codeframe}{HTML}{CCCCCC}
\definecolor{cprimary}{HTML}{2166AC}
\definecolor{csecondary}{HTML}{B2182B}
\definecolor{ctertiary}{HTML}{2CA02C}
\tcbset{
  leanbase/.style={
    listing only,
    breakable,
    listing options={language=lean, numbers=none,
      basicstyle=\ttfamily\footnotesize, backgroundcolor=\color{white},
      aboveskip=0pt, belowskip=0pt},
    colback=white,
    boxrule=0pt,
    leftrule=2pt,
    arc=0pt,
    left=4pt, right=2pt, top=2pt, bottom=2pt,
  },
}
\newtcblisting{leancode}{
  leanbase,
  colframe=cprimary,
  before skip=0.8em,
  after skip=0.5em,
}
\newtcblisting{leancode-blue}{
  leanbase,
  colframe=cprimary,
  colback=cprimary!5,
  listing options={language=lean, numbers=none,
    basicstyle=\ttfamily\footnotesize, backgroundcolor=\color{cprimary!5},
    aboveskip=0pt, belowskip=0pt},
  before skip=2pt,
  after skip=2pt,
}
\newtcblisting{leancode-green}{
  leanbase,
  colframe=ctertiary,
  colback=ctertiary!5,
  listing options={language=lean, numbers=none,
    basicstyle=\ttfamily\footnotesize, backgroundcolor=\color{ctertiary!5},
    aboveskip=0pt, belowskip=0pt},
  before skip=2pt,
  after skip=2pt,
}
\newtcblisting{leancode-red}{
  leanbase,
  colframe=csecondary,
  colback=csecondary!5,
  listing options={language=lean, numbers=none,
    basicstyle=\ttfamily\footnotesize, backgroundcolor=\color{csecondary!5},
    aboveskip=0pt, belowskip=0pt},
  before skip=2pt,
  after skip=2pt,
}

\def\framework{P$^3$}

\title{P$^{3}$: Joint Program-and-Proof Planning\\
  for Verified Code Generation}

\author{%
  Zenan Li$^{1,*}$ \quad
  Ziran Yang$^{2,*}$ \quad
  Peiyang Song$^{3}$ \quad
  Zhaoyu Li$^{4}$ \quad
  Kaiyu Yang$^{1}$\\[0.5em]
  $^{1}$Apodex \quad
  $^{2}$Princeton University \quad
  $^{3}$Caltech \quad
  $^{4}$University of Toronto\\[0.25em]
  $^{*}$Equal contribution
}

\hypersetup{
  pdftitle={P3: Joint Program-and-Proof Planning for Verified Code Generation},
  pdfauthor={Zenan Li, Ziran Yang, Peiyang Song, Zhaoyu Li, Kaiyu Yang},
}

\begin{document}

\maketitle

\begin{abstract}
Verified code generation asks a large language model (LLM) to generate both an executable program and a machine-checkable proof that the program meets a formal specification, promising software that is correct by construction.
The de facto workflow decouples the two halves of the problem: first synthesize a program, then attempt to prove it correct. We observe that this sequential pipeline can be both ineffective and inefficient in practice. A program generated without anticipating its proof can be subtly incorrect or structurally difficult to verify, forcing the LLM into brittle repair loops that alternate between patching the code and patching the proof. Inspired by Dijkstra's view that a program and its correctness argument should be developed hand in hand, we propose \framework{}, an LLM-based agentic workflow that first derives a unified program-and-proof plan from the specification, then elaborates the implementation and proof scaffold under this shared plan. To evaluate verified code generation in realistic settings, we further introduce Lean4Commit0, a repository-derived, library-level benchmark built by extracting core APIs from real-world software repositories and translating their requirements, including relational specifications across APIs, into Lean tasks.
Using four frontier LLM backends, we evaluate \framework{} on Verina, AlgoVeri, and our Lean4Commit0 benchmark, where it achieves the highest solve rate in every benchmark--model setting.
Compared with the stronger baseline, it improves solve rates by 4.6--11.2 percentage points and reduces per-task API cost by up to roughly 40\% and wall-clock time by up to roughly 37\% on the difficult subset of each benchmark.
A targeted ablation further shows gains of 3.3--8.3 points over implementation-only planning, isolating the benefit of planning the program and proof jointly.
\end{abstract}

\section{Introduction}
\label{sec:intro}

Recent progress in large language models (LLMs) has pushed code generation beyond autocomplete-style snippets to full functions, files, and repository-level development through agentic workflows~\citep{liu2024large, dong2025survey, jiang2026survey}. At the same time, formal systems such as Lean~4~\citep{moura2021lean} have become practical environments for writing executable programs together with machine-checkable proofs of their correctness~\citep{baanen2025growing, barrett2026cslib}.
The intersection of these two trends gives rise to \emph{verified code generation}~\citep{ye2025verina, dougherty2025proving}: given a formal specification, an LLM is asked to synthesize both an executable implementation and a proof that the implementation satisfies the specification. If the formal system accepts both artifacts, the generated program carries a machine-checked correctness guarantee that goes beyond finite test coverage, substantially reducing the burden of reviewing, trusting, and composing LLM-written software.

Despite this promise, existing verified-code-generation pipelines often follow a sequential \emph{program-then-proof} schedule: the LLM first commits to a concrete implementation, after which a separate proving stage, ranging from automated solvers to LLM-based provers, is invoked to discharge the resulting proof obligations~\citep{gloeckle2026wybecoder,li2026goedel,yang2025autoverus,yang2025verusage,yang2026exverus}.
When verification fails, the workflow enters a patch-and-retry loop.
This ordering creates both \emph{effectiveness} and \emph{efficiency} bottlenecks. First, an implementation generated without anticipating its proof may lack the invariant or induction structure needed for a clean correctness argument. In such cases, local repair is unlikely to recover a proof-friendly structure that was never planned. Second, even when a verified solution is eventually found, much of the token and time budget may be spent exploring repairs that are doomed by the initially committed program.

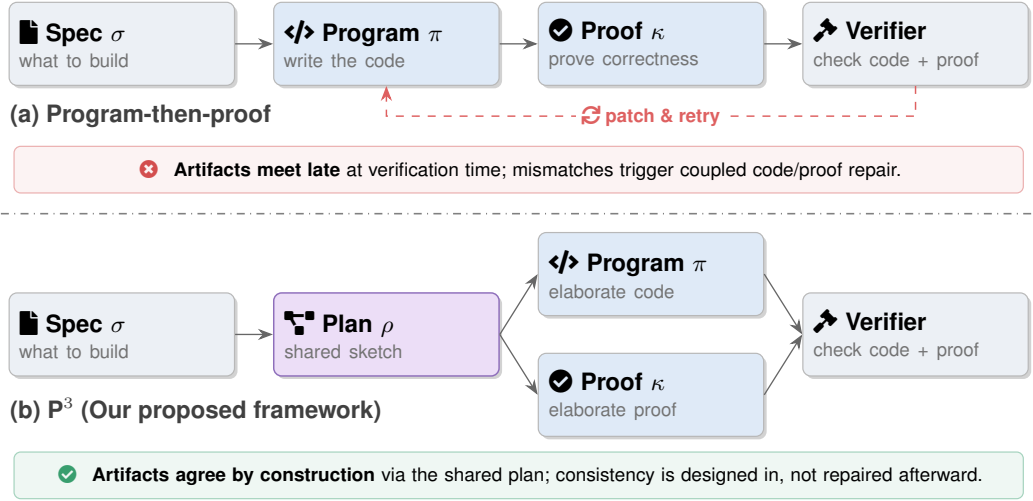
\begin{figure}[t]
  \centering

\providecommand{\fwsub}{\scriptsize\color{black!55}}

\definecolor{fwSpecFill}{RGB}{236, 240, 245}
\definecolor{fwBoxFill}{RGB}{223, 234, 247}
\definecolor{fwPlanFill}{RGB}{236, 222, 247}
\definecolor{fwPlanBorder}{RGB}{142, 110, 187}
\definecolor{fwRed}{RGB}{211, 78, 78}
\definecolor{fwGreen}{RGB}{56, 158, 110}
\definecolor{fwLoopRed}{RGB}{222, 96, 96}
\definecolor{fwLoopPurple}{RGB}{142, 110, 187}

\tikzset{
  fwbox/.style={
    rectangle, rounded corners=3pt, draw=black!28, line width=0.45pt,
    align=left, inner sep=4pt, text width=27mm, minimum height=11mm,
    font=\small\sffamily,
    drop shadow={shadow xshift=0.4mm, shadow yshift=-0.4mm, opacity=0.18, fill=black}
  },
  fwspec/.style={fwbox, fill=fwSpecFill},
  fwbody/.style={fwbox, fill=fwBoxFill},
  fwplan/.style={fwbox, fill=fwPlanFill, draw=fwPlanBorder, line width=0.55pt},
  fwarr/.style={-{Stealth[length=2mm]}, line width=0.5pt, draw=black!60},
  fwloop/.style={-{Stealth[length=2mm]}, line width=0.55pt, dashed, draw=fwLoopRed},
  fwretreat/.style={-{Stealth[length=2mm]}, line width=0.55pt, dashed, draw=fwLoopPurple},
  fwttl/.style={font=\small\sffamily\bfseries, color=black!75, anchor=west},
  fwlbl/.style={font=\scriptsize\sffamily, fill=white, inner sep=2pt, align=center},
  fwnoteR/.style={
    rectangle, rounded corners=2.5pt, draw=fwRed!45, fill=fwRed!7, line width=0.45pt,
    align=center, inner sep=5pt, font=\scriptsize\sffamily, text width=130mm
  },
  fwnoteG/.style={
    rectangle, rounded corners=2.5pt, draw=fwGreen!45, fill=fwGreen!8, line width=0.45pt,
    align=center, inner sep=5pt, font=\scriptsize\sffamily, text width=130mm
  },
}

\begin{tikzpicture}
  \node[fwspec] at (1.6, -1.05) (specA)
    {\faFile~\textbf{Spec} $\sigma$\\\fwsub what to build};
  \node[fwbody] at (5.1, -1.05) (progA)
    {\faCode~\textbf{Program} $\pi$\\\fwsub write the code};
  \node[fwbody] at (8.6, -1.05) (proofA)
    {\faCheckCircle~\textbf{Proof} $\kappa$\\\fwsub prove correctness};
  \node[fwspec] at (12.1, -1.05) (verA)
    {\faGavel~\textbf{Verifier}\\\fwsub check code + proof};

  \draw[fwarr] (specA)  -- (progA);
  \draw[fwarr] (progA)  -- (proofA);
  \draw[fwarr] (proofA) -- (verA);

  \draw[fwloop] (verA.south) -- ++(0,-4mm) -| (progA.south);
  \node[fwlbl, text=fwLoopRed] at (8.6, -2)
    {\faSync\ \textbf{patch \& retry}};

  \node[fwttl] at (0, -2.0) (titA) {(a) Program-then-proof};

  \node[fwnoteR, anchor=north] at (6.85, -2.4)
    {\textcolor{fwRed}{\faTimesCircle}\ \,
     \textbf{Artifacts meet late} at verification time;
     mismatches trigger coupled code/proof repair.};

  \draw[dash dot, line width=0.6pt, color=black!50] (0, -3.3) -- (13.7, -3.3);

  \node[fwspec] at (1.6, -4.9) (specB)
    {\faFile~\textbf{Spec} $\sigma$\\\fwsub what to build};
  \node[fwplan] at (5.1, -4.9) (planB)
    {\faProjectDiagram~\textbf{Plan} $\rho$\\\fwsub shared sketch};
  \node[fwbody] at (8.6, -4.1) (progB)
    {\faCode~\textbf{Program} $\pi$\\\fwsub elaborate code};
  \node[fwbody] at (8.6, -5.7) (proofB)
    {\faCheckCircle~\textbf{Proof} $\kappa$\\\fwsub elaborate proof};
  \node[fwspec] at (12.1, -4.9) (verB)
    {\faGavel~\textbf{Verifier}\\\fwsub check code + proof};

  \draw[fwarr] (specB)        -- (planB);
  \draw[fwarr] (planB.east)   -- (progB.west);
  \draw[fwarr] (planB.east)   -- (proofB.west);
  \draw[fwarr] (progB.east)   -- (verB.west);
  \draw[fwarr] (proofB.east)  -- (verB.west);

  \node[fwttl] at (0, -5.9) (titB) {(b) \framework{} (Our proposed framework)};

  \node[fwnoteG, anchor=north] at (6.85, -6.45)
    {\textcolor{fwGreen}{\faCheckCircle}\ \,
     \textbf{Artifacts agree by construction} via the shared plan;
     consistency is designed in, not repaired afterward.};
\end{tikzpicture}
  \caption{Our proposed framework. \textit{Top:} the program-then-proof pipeline commits to an implementation before the proof's structure is known, so the two artifacts only meet at verification time and mismatches trigger coupled code/proof repair. \textit{Bottom:} our workflow first settles a shared structural plan, then elaborates the program and proof scaffold against the same plan, so that consistency holds by construction rather than being recovered by post-hoc repair.}
  \label{fig:framework}
  \vspace{-10pt}
\end{figure}

To address these limitations, we propose \framework{}, a \emph{joint \underline{p}rogram-and-\underline{p}roof \underline{p}lanning} framework for verified code generation in Lean~4, ensuring that the implementation and proof are structurally consistent before either artifact is elaborated.
The underlying principle is classical. Programs and their correctness arguments have long been viewed as co-designed artifacts: \citet{dijkstra1976discipline} argued that proof structure and program structure mutually constrain each other, and therefore that the two should be developed ``hand in hand''.
We adapt this principle to an agentic setting by having the model first construct a shared high-level plan for both artifacts. This plan specifies the decomposition that carries the computation, the intermediate assertions or invariants that connect the precondition to the postcondition, the induction or case analysis that structures the proof, and the auxiliary lemmas required along the way.
The implementation and proof scaffold are then elaborated against this same plan, with remaining proof leaves discharged using Lean~4's feedback. As a result, the two artifacts agree structurally from the start, rather than being reconciled through post-hoc repair (Figure~\ref{fig:framework}).

To evaluate verified code generation beyond textbook settings, we also introduce Lean4Commit0, a benchmark derived from real open-source software. Existing benchmarks for verified code generation primarily consist of textbook algorithms and competitive-programming tasks, which are valuable but often differ from the APIs and behaviors encountered in production software. Lean4Commit0 is built from 108 real-world open-source libraries spanning four mainstream ecosystems: Python, Rust, C/C++, and Java.
For each library, we select several core APIs and encode their expected behavior as Lean implementation placeholders and correctness theorems, including relational specifications that connect multiple API calls.
To ensure specification quality, each specification passes a three-stage validation pipeline: a \emph{reference-satisfaction} check verifies that the behavior of the original implementation is consistent with the Lean specification; a \emph{mutation-rejection} check tests whether the specification rejects behavior-changing mutants; an \emph{LLM review} flags specifications that are trivial or irrelevant.

We evaluate \framework{} on Verina~\citep{ye2025verina}, AlgoVeri~\citep{zhao2026algoveri}, and Lean4Commit0 with four frontier LLM backends, comparing against a plain Lean-agent baseline and a program-then-proof baseline. Across all 12 (benchmark, model) cells, \framework{} achieves the highest solve rate, with a 4.6--11.2 percentage-point gain over the stronger primary baseline in each setting; on difficult tasks, it also reduces per-task cost and time by up to 40\% and 37\%, respectively. A targeted ablation with implementation-only planning yields a further controlled comparison: joint program-and-proof planning improves solve rate by 3.3--8.3 points.
A case study on left-leaning red-black tree deletion further shows that the program-then-proof baseline either fails after 6,344 lines under the structural-deletion commitment or succeeds only with redundant rebuild invariants in 1,176 lines, whereas \framework{} compares the plans before committing and solves the task in 1,105 lines.



\section{Problem}
\label{sec:problem}

The input to \emph{verified code generation} is a formal specification $\sigma = (P, Q)$ for a target function, where $P$ is a precondition on the function's inputs and $Q$ is a postcondition relating those inputs to the outputs. Given $\sigma$, the goal is to construct two artifacts: a \emph{program} $\pi$ implementing the function, and a \emph{certificate} $\kappa$ from which a verifier can establish that, on every input satisfying $P$, running $\pi$ terminates and returns an output satisfying $Q$. A task is solved iff a designated verifier accepts $(\sigma, \pi, \kappa)$.
This setting differs from test-based code generation in that correctness must be established for \emph{all} admissible inputs rather than for a finite collection of examples.
The concrete form of $\kappa$ depends on the verification backend. In SMT-based systems, such as Dafny~\citep{leino2010dafny}, F$^\star$~\citep{swamy2016dependent}, and Verus~\citep{lattuada2023verus}, $\kappa$ is a collection of program annotations whose verification conditions are discharged by an SMT solver. In interactive theorem proving settings, such as Lean~4~\citep{moura2021lean}, Rocq~\citep{bertot2004coq}, and Isabelle/HOL~\citep{nipkow2002isabelle}, $\kappa$ is an explicit proof term or tactic script checked by the prover's kernel.

Although these backends expose different certificate languages, representative LLM agents largely follow the same sequential schedule. In Verus, systems such as AutoVerus~\citep{yang2025autoverus}, VeruSAGE~\citep{yang2025verusage}, and ExVerus~\citep{yang2026exverus} first commit to a program and then repair its annotations; Lean~4 systems such as WybeCoder~\citep{gloeckle2026wybecoder} and Goedel-Code-Prover~\citep{li2026goedel} follow the analogous loop over tactic scripts.
Across both settings, the LLM agent first commits to an implementation, then searches for a compatible certificate, using verifier feedback to patch the certificate and, when necessary, revise the program.

The limitation is that the implementation largely determines the shape of the certificate. A structurally inconvenient program can force the LLM to establish auxiliary lemmas that are not directly suggested by the specification, making certificate search unnecessarily difficult. This coupling between code structure and proof structure is especially visible in Lean~4 benchmarks such as Verina~\citep{ye2025verina} and AlgoVeri~\citep{zhao2026algoveri}, where each task provides a function signature together with pre- and postconditions, and leaves both the implementation and proof as open placeholders.

As a running example, consider the task of computing the maximum element of a non-empty list of integers, together with a theorem stating that the returned value belongs to the list and upper-bounds every element. Two natural implementations are possible (see Appendix~\ref{app:listmax} for the full Lean source of both versions and their proofs):
\vspace{-0.35em}

\begin{minipage}[t]{0.56\linewidth}
\begin{leancode-green}
-- (A) structural recursion
def listMax : (xs : List Int) → xs ≠ [] → Int
  | [x],     _ => x
  | x::y::t, _ => max x (listMax (y::t) (by simp))
\end{leancode-green}
\end{minipage}%
\hfill
\begin{minipage}[t]{0.4\linewidth}
\begin{leancode-green}
-- (B) accumulator fold
def listMax (xs : List Int)
    (h : xs ≠ []) : Int :=
  xs.foldl max xs.head!
\end{leancode-green}
\end{minipage}
\vspace{-0.35em}

Both implementations compute the same value, but they induce very different proof obligations. Version~(A) admits an induction that mirrors the postcondition: membership and upper-bound facts are proved case by case, with the induction hypothesis applied directly to the tail. Version~(B), although shorter and arguably more idiomatic, is in fact substantially harder to verify against the stated specification. A proof cannot proceed directly from the original postcondition; it must first generalize to an accumulator invariant such as
\vspace{-0.25em}

\begin{minipage}[t]{\linewidth}
\begin{leancode-blue}
lemma foldl_max_gen (init : Int) (ys : List Int) :
    ys.foldl max init ∈ init :: ys ∧ ∀ z ∈ init :: ys, z ≤ ys.foldl max init
\end{leancode-blue}
\end{minipage}
\vspace{-0.35em}
\noindent


This lemma is then specialized to the non-empty input \lean{xs} to recover the desired theorem. The universal quantification over arbitrary \lean{init} has no counterpart in the specification: it is an invariant of \emph{the fold}, not of \emph{the maximum} itself. Thus, after committing to Version~(B), the LLM agent must either discover this stronger invariant under the fixed implementation or rewrite the program toward Version~(A), discarding any proof work already accumulated. The failure mode is therefore structural: the sequential schedule commits the implementation before identifying the proof structure most natural for the specification, often leading to wasted certificate search.

\section{Methodology}
\label{sec:method}

\textbf{Motivation.}
The central limitation of the program-then-proof paradigm is the potential misalignment between program and proof construction. A program may satisfy the specification extensionally, yet expose a structure that is difficult to verify.
Conversely, optimizing only for proof simplicity can produce implementations that are inefficient, unnatural, or fail to capture the intended algorithm.
For example, consider searching for a target in a sorted array.
Binary search has the desired asymptotic behavior, but its proof requires reasoning about intervals, index bounds, and preservation of the search invariant across recursive calls.
A linear scan is much easier to verify by induction, but it ignores the sortedness assumption and gives a weaker algorithmic result.
This tension motivates planning the program and proof jointly, rather than generating one first and forcing the other to fit afterward.

\textbf{Framework.} Given a formal specification $\sigma = (P, Q)$, our LLM agent, \framework{}, runs in two stages.
(1)~\emph{Planning:}
\framework{} proposes candidate plans based on $\sigma$ and selects a plan $\rho$ containing a program sketch aligned with a compatible proof sketch.
(2)~\emph{Elaboration:} conditioned on $\rho$, \framework{} synthesizes the program $\pi$ and a proof scaffold that mirrors $\rho$ (e.g., induction principles, case splits, and intermediate lemmas).
It then fills in the remaining proof details, using Lean feedback for repair.
On success, Lean certifies the proof $\kappa$ of correctness for $(\sigma, \pi)$; on failure, \framework{} classifies the error as elaboration-level or plan-level and invokes the corresponding repair procedure.
Within a single agent harness, planning and elaboration run as two stages with distinct prompts and tool contexts; the main agent (running the named backend) handles planning, and for elaboration may either continue itself or dispatch cheaper subagents. The concrete templates are given in Appendix~\ref{app:prompts}.


\subsection{Joint Program-and-Proof Planning}
\label{sec:method:planning}


A plan $\rho$ records four commitments that the program and proof must later realize: (i)~the formal contract and any algorithmic or asymptotic instruction; (ii)~the program decomposition, including the recursion variable, branch structure, data representation, auxiliary routines, and termination measure; (iii)~the concrete library functions and existing lemmas on which elaboration will rely; and (iv)~the proof obligations, including the bridging predicate, induction or case structure, and statements of auxiliary lemmas.
Together, these choices specify the structural shape that the program and proof must share.

The planning agent validates the candidate plan at the sketch level.
This validation is informal but explicit: the program sketch must meet the task instructions, including any algorithmic constraints; the proof sketch must explain why the chosen decomposition preserves the bridging predicate and why that predicate entails $Q$; and any nontrivial strengthening of $Q$ needed by the proof must be named as an auxiliary lemma rather than left implicit.
Only after this validation does the agent pass the selected plan to elaboration.

\noindent\textbf{Compact procedure.}
\begin{enumerate}[leftmargin=*,topsep=2pt,itemsep=0pt,parsep=0pt]
  \item Propose candidate plans from $\sigma$, each containing all four commitments above.
  \item Reject a candidate if its program sketch violates the instruction, its decomposition does not preserve the bridging predicate, or the predicate does not entail $Q$.
  \item Select a surviving candidate whose decomposition exposes the smallest anticipated proof burden, and freeze its structural commitments as $\rho$.
  \item Elaborate $(\pi,\kappa)$ under $\rho$; repair local errors in place, but return to Step~1 if elaboration requires an unplanned algorithm, invariant, or structural lemma.
\end{enumerate}

In the running \lean{listMax} example, planning exposes the structural tradeoff between Version~(A) and Version~(B) before the agent commits to code.
Version~(A) uses structural recursion on \lean{xs} (split \lean{[x]} vs.\ \lean{x :: y :: t}) and lets the postcondition itself serve as the bridging predicate.
Version~(B) takes \lean{List.foldl}, which forces the bridging predicate to range over a universally quantified accumulator, stated explicitly as the lemma \lean{foldl_max_gen}.
This generalized accumulator invariant is not a surface fact about the maximum function; it is a proof obligation introduced by the choice to implement the function with \lean{List.foldl}.

If the fold choice is made only as code, the agent sees the invariant only after proof search fails.
If the same choice is represented in the plan, the invariant appears immediately as part of the proof sketch, before any tactic script is written.
The planner can therefore compare version~(A) and version~(B) at the level of sketches: version~(A) aligns the computation with the postcondition directly, whereas version~(B) commits the agent to proving and later specializing an auxiliary accumulator lemma.
This comparison is lightweight because plans differ only in their decomposition, bridging predicate, and named auxiliary lemmas, not in full Lean terms or tactic scripts.
Although \lean{listMax} is our running example, the same kind of sketch-level comparison applies to other structural choices as well, including case splits, collection combinators, and relational specifications across multiple functions.

\subsection{Plan-Guided Elaboration}
\label{sec:method:elaboration}

Once a plan $\rho$ is selected, elaboration turns its sketches into concrete Lean artifacts.
For programs, the agent writes an implementation whose control structure follows the decomposition fixed by $\rho$: the recursive argument, case split, collection combinator, or auxiliary routine is not chosen again during coding.
For proofs, the agent writes a certificate scaffold with the same shape: the target theorem, the auxiliary lemma statements named by the plan, and the top-level proof structure that invokes the planned induction, case analysis, or lemma applications.
At this point the program is concrete, while the scaffold may still contain local gaps, represented as \lean{sorry}, at the leaves of the planned argument.

Proof completion then becomes a local elaboration problem rather than an unconstrained proof search problem.
The agent asks Lean to elaborate the file, reads the resulting goals and errors, and fills the remaining proof leaves or auxiliary lemmas while preserving the structure fixed by $\rho$.
For example, if $\rho$ specifies structural recursion over a list, the agent closes the cases exposed by that recursion; if $\rho$ specifies a fold with an accumulator invariant, the agent works on the accumulator lemma and its specialization rather than inventing a different induction principle inside the main proof.

This completion step can be implemented by proof-tree search, test-time scaling over candidate tactics, or an agentic Lean-feedback loop.
In our Lean~4 setting, LSP/MCP feedback provides localized goals and diagnostics without rebuilding the search context from scratch.
The verifier still has the final authority: the task is solved only when Lean accepts the full pair $(\pi,\kappa)$ against $\sigma$.

\subsection{Benefits of Plan-Guided Generation}
\label{sec:method:benefits}

The plan turns the verifier's binary feedback into structured guidance about \emph{where} a failure lives and \emph{what} progress can be salvaged.
An \emph{elaboration-level} failure is one whose error localizes under the current plan, such as a missing tactic in one branch, an unmet preservation step for the bridging predicate, or a small implementation mismatch in $\pi$.
The agent repairs the localized artifact while keeping the rest of the plan and the other side of the pair intact.
A \emph{plan-level} failure is signaled when repeated local repairs cannot close the obligations under $\rho$, or when Lean exposes that the bridging predicate is too weak, for example because completion requires an algorithm, invariant, or structural lemma that the plan did not name.
In that case the agent retreats to planning, revises $\rho$, and only then re-elaborates the program and certificate against the new plan.
Thus the plan serves as a single, well-defined point of retreat: when nothing under it works, the agent moves up one level rather than thrashing between code edits and proof edits, which is the failure mode that drives the efficiency gap of program-then-proof pipelines.

This planning-and-elaboration separation also changes the model-allocation problem.
Conceptually, the high-level reasoning load is concentrated in planning, where the agent must choose the decomposition, identify the bridging predicate, and decide which auxiliary lemmas make the code sketch and proof sketch compatible.
Once those choices are fixed, elaboration is a more constrained implementation task: the agent is asked to realize an agreed-upon structure in Lean, not to rediscover the structure from verifier failures.
In practice, the main agent uses the named backend for planning, and for elaboration steps that it deems constrained enough it may dispatch cheaper subagents rather than handle them itself.
This division is not part of the trusted base: Lean still checks the final artifact, and an elaborator can only succeed by producing code and proofs accepted by the kernel.

\section{Benchmark}
\label{sec:benchmark}

Lean4Commit0 extends the single-function setting to library tasks in which several APIs are jointly constrained by relational postconditions.
Existing Lean~4 benchmarks for verified code generation, including Verina~\citep{ye2025verina}, Clever~\citep{thakur2025clever}, and AlgoVeri~\citep{zhao2026algoveri}, draw most of their tasks from textbooks and competitive programming.
These tasks are typically short, self-contained, and centered on a single function whose behavior is captured by one or two arithmetic or list invariants.
Real software, however, has a different texture: state that is shared across several APIs, ordering and consistency properties that relate multiple operations, and edge cases that arise from actual usage rather than from a one-line problem statement.
Without benchmark coverage of these patterns, claims about verified code generation remain hard to interpret for the kinds of code developers actually ship.
To close this gap, and inspired by Commit0~\citep{zhao2024commit0}, we present \emph{Lean4Commit0}: a benchmark that extracts Lean~4 verified-generation tasks from 108 real-world open-source libraries.

\subsection{Benchmark Construction}
\label{sec:benchmark:construction}

The source libraries span four ecosystems: 65 Python, 17 Rust, 15 C/C++, and 11 Java.
They cover a wide range of domains, including cryptographic primitives, data structures, web and network frameworks, parsers, schedulers, and game emulators.
From each library we select several \emph{core APIs}, defined as the user-facing entry points to that library's main data structures. Each selected API is encoded as a triple of (i)~a Lean function signature with body \lean{sorry}, (ii)~a postcondition predicate, and (iii)~a correctness theorem with proof \lean{sorry}. When the intended semantics ties several APIs together, we additionally add \emph{relational specs}: theorems whose statement quantifies over the joint behavior of two or more functions. Each library forms one task; across the 108 libraries, the benchmark contains 511 implementation placeholders and 519 theorem placeholders.

For example, Fabric is a Python library for SSH-based deployment and remote execution; Lean4Commit0 models core configuration logic from its API.
We extract operations such as \lean{Config.set} and \lean{Config.get}, where configuration entries carry precedence levels from defaults to runtime overrides. Beyond single-API behavior, Lean4Commit0 adds relational specifications that connect operations across calls: after setting key \lean{k} to value \lean{v}, a subsequent lookup of \lean{k} returns \lean{some v}; and if the same key is set at a lower level and then a higher level, \lean{get} returns the higher-precedence value. These properties cannot be discharged by reasoning about \lean{set} or \lean{get} in isolation. The proof must connect updates to later queries and respect the precedence ordering. The full encoded Lean task appears in Appendix~\ref{app:fabric}.


\subsection{Specification Quality Pipeline}
\label{sec:benchmark:quality}

Repository-grounded specifications must be adequate with respect to the intended library behavior, but this adequacy cannot be checked directly: for real libraries, the intended behavior is distributed across source code, tests, documentation, and usage conventions rather than given as a complete formal oracle~\citep{zowghi2003interplay,montgomery2022empirical}.
Moreover, the available executable evidence, such as source code and tests, lives in the original project language rather than in Lean, preventing direct Lean-level adequacy checking.
Therefore, Lean4Commit0 uses an automated filtering pipeline to detect three common and actionable failure modes:
\begin{itemize}[leftmargin=*,itemsep=0pt,topsep=2pt]
  \item \emph{Unsatisfiable}: no implementation can satisfy the stated theorem obligations.
  \item \emph{Underconstrained}: behavior-changing implementations can still satisfy the stated properties.
  \item \emph{Vacuous}: the stated properties are trivial or irrelevant to the target API behavior.
\end{itemize}

To detect unsatisfiable specifications, we use the reference implementation and test cases from the original project as executable evidence.
For each test input, we run the reference implementation to obtain the reference output, then check in Lean whether the resulting input-output pair satisfies the corresponding pre/postcondition or relational specification.
To detect underconstrained specifications, we apply behavior-changing mutations to the reference implementation and run the same test inputs again.
The mutated input-output pairs should violate the specification; if they still satisfy it, the spec is too weak to distinguish the mutant from the reference behavior.
Formal counterexample generation likewise uses symbolic mutation to construct Lean-checkable negative instances~\citep{li2026disprove}; our quality gate instead mutates executable library implementations and asks whether candidate API specifications reject their changed behavior.

To detect vacuous specifications, we invoke an LLM-based review with explicit rules: the specification should not be provable by trivial tactics such as \lean{rfl} or \lean{decide}, should not be a tautology or duplicate of another statement, and should depend on the functions being implemented.
The LLM reviewer finally assigns a score on a 1--5 scale based on these criteria and flags specifications that add little or no constraint on the target API behavior.

To automate the specification quality-assurance pipeline end to end, Lean4Commit0 introduces an LLM agent that orchestrates the components above.
For example, we provide various mutation tools for the supported source languages; the agent selects and invokes the appropriate tool for each project to obtain behavior-changing mutants.
The agent also bridges the language gap between the reference project and Lean: since the reference projects are written in Python, Rust, C/C++, or Java while the specifications are written in Lean, it generates language-specific shims that run the reference code, collect inputs and outputs, and serialize them as Lean values.

\begin{table}[t]
  \centering
  \caption{Lean4Commit0 summary after the specification quality pipeline. Reference satisfaction is a hard admissibility condition; the quality score combines mutation rejection and LLM review.}
  \label{tab:lean4commit0-summary}
  \small
  \begin{tabular}{lrrrrrrr}
    \toprule
    \multirow{2.5}{*}{Language} & \multirow{2.5}{*}{Libraries} & \multicolumn{3}{c}{Placeholders (\#)} & \multicolumn{3}{c}{Quality metrics (\%)} \\
    \cmidrule(lr){3-5}\cmidrule(lr){6-8}
    & & Code & Theorem & Total & Mutation & Review & Quality \\
    \midrule
    Python & 65 & 314 & 314 & 628 & 98.7 & 82.5 & 90.6 \\
    Rust & 17 & 88 & 73 & 161 & 98.2 & 86.2 & 92.2 \\
    C/C++ & 15 & 58 & 72 & 130 & 98.7 & 80.9 & 89.8 \\
    Java & 11 & 51 & 60 & 111 & 99.0 & 80.3 & 89.7 \\
    \midrule
    All & 108 & 511 & 519 & 1{,}030 & 98.6 & 82.7 & 90.7 \\
    \bottomrule
  \end{tabular}
  \vspace{-10pt}
\end{table}

We treat reference satisfaction as a hard admissibility condition: every reference input-output pair must satisfy the corresponding Lean specification.
For specifications that pass this check, we compute an aggregate quality score from mutation rejection and LLM review.
Let $\mathrm{Mutation}$ denote the mutation rejection rate, i.e., the fraction of behavior-changing mutants rejected by the specification, and let $\mathrm{Review}$ denote the LLM review score for vacuity and semantic relevance, normalized to a percentage as $\mathrm{Review}(\sigma) = \mathrm{score}(\sigma)/5 \times 100\%$.
We define
\begin{equation*}
  \mathrm{Quality}(\sigma)
  = 0.5 \cdot \mathrm{Mutation}(\sigma)
  + 0.5 \cdot \mathrm{Review}(\sigma).
\end{equation*}
We iteratively refine each task and admit it only if reference satisfaction is complete and the aggregate score is at least 80\%.
Table~\ref{tab:lean4commit0-summary} summarizes the resulting benchmark scale and specification-quality metrics after this pipeline.
All 108 libraries clear this threshold; the mean score across the benchmark is 90.7\%, and ten libraries reach 100\%.

\section{Experiments}
\label{sec:experiments}

\subsection{Experimental Setup}
\label{sec:experiments:setup}

We evaluate whether the proposed framework improves verified code generation along two axes: \emph{effectiveness}, measured by how often the agent produces a Lean-accepted implementation and proof, and \emph{efficiency}, measured by how much repair work is needed before acceptance.

\textbf{Benchmarks.} We evaluate on Verina~\citep{ye2025verina} (189 tasks), AlgoVeri~\citep{zhao2026algoveri} (77 tasks), as well as the new Lean4Commit0 benchmark.
Verina and AlgoVeri test single-function verified generation over textbook-style algorithmic tasks, while Lean4Commit0 tests repository-derived, library-level task bundles, including relational specifications across APIs.
Because Verina and AlgoVeri do not impose explicit complexity requirements, agents can sometimes satisfy the formal specification with impractical implementations, such as exponential-time algorithms.
We therefore augment each task with a natural-language instruction specifying the expected algorithmic strategy or asymptotic complexity, and apply the same instruction to all evaluated systems.

\textbf{Baselines.}
We compare the proposed framework against two primary agent settings.
Existing proof-search and annotation-repair systems assume a fixed implementation, so they do not provide a direct end-to-end orchestration baseline (Section~\ref{sec:related}).
The first is a plain Lean-agent setting: the agent receives the formal specification and Lean tools~\citep{lean-lsp-mcp}, but no task-specific skill or strategy prompt.
The second is a skill-based program-then-proof setting: the agent is guided to first produce an implementation and then generate and repair the proof~\citep{lean4-skills}.
Our framework instead equips the agent with a planning skill: before elaboration, it selects a plan $\rho$ containing a mutually consistent program sketch and proof sketch, and then writes both artifacts under that plan.
For a targeted ablation, we additionally evaluate \textsc{Plan-Seq}, which plans implementation-side choices but is forbidden from anticipating invariants, proof decomposition, induction strategies, or proof lemmas before running the program-then-proof pipeline.

\textbf{Settings.}
We instantiate each agent setting with four frontier backend configurations: Codex-GPT-5.5, Gemini-3-Pro, Claude-Sonnet-4.6, and Claude-Opus-4.7, invoked through their native command-line interfaces under a fixed harness.
Each run is given a per-task API budget of 30.0 USD and a wall-clock time limit of 180 minutes.
All runs execute inside the same Docker image with Lean \texttt{4.28.0}.
To isolate the effect of the agent setting, all runs use the same task instructions, API budget, time limit, and maximum feedback-iteration budget.
Each $(\text{task},\text{model},\text{method})$ configuration is run once, so rates aggregate over tasks but do not estimate within-task variance; the complete prompts, harness, scripts, pinned environment, and run metadata are available at \url{https://figshare.com/s/2ab7020cacaffcf0d258}.

\subsection{Main Results}
\label{sec:experiments:main}

\begin{table}[t]
  \centering
  \caption{Solve rate (\%) across backend models, methods, and benchmarks. A task is counted as solved if the produced solution satisfies the instruction and passes Lean's checker. Higher is better.}
  \label{tab:main-solve}
  \small
  \begin{tabular}{>{\raggedright\arraybackslash}p{2.8cm}l *{3}{>{\raggedleft\arraybackslash}p{1.4cm}} >{\columncolor{black!10}\raggedleft\arraybackslash}p{1.4cm}}
    \toprule
    Benchmark & Model & \textsc{Plain} & \textsc{Seq} & \framework{} & $\Delta$ \\
    \midrule
    \multirow{4}{*}{Verina}
      & Codex-GPT-5.5       & 72.0    & 67.7    &  \textbf{77.2}    & +5.2 \\
      & Gemini-3-Pro      & 59.3    & 60.8    & \textbf{72.0}    & +11.2 \\
      & Claude-Sonnet-4.6 & 64.0 & 61.9 & \textbf{73.0} & +9.0 \\
      & Claude-Opus-4.7   & 68.8 & 68.3 & \textbf{74.6} & +5.8 \\
    \midrule
    \multirow{4}{*}{AlgoVeri}
      & Codex-GPT-5.5       & 36.4 & 33.8 & \textbf{44.2} & +7.8 \\
      & Gemini-3-Pro      & 33.8 & 31.2 & \textbf{39.0} & +5.2 \\
      & Claude-Sonnet-4.6 & 35.1 & 32.5 & \textbf{40.3} & +5.2 \\
      & Claude-Opus-4.7   & 39.0 & 40.3 & \textbf{48.1} & +7.8 \\
    \midrule
    \multirow{4}{*}{Lean4Commit0}
      & Codex-GPT-5.5       & 11.1 & 13.0 & \textbf{18.5} & +5.5 \\
      & Gemini-3-Pro      & \phantom{0}9.3 & 11.1 & \textbf{15.7} & +4.6 \\
      & Claude-Sonnet-4.6 & 10.2 & 13.0 & \textbf{17.6} & +4.6 \\
      & Claude-Opus-4.7   & 13.9 & 17.6 & \textbf{22.2} & +4.6 \\
    \bottomrule
  \end{tabular}
  \vspace{-10pt}
\end{table}

Table~\ref{tab:main-solve} reports solve rates, where a task counts as solved only when the produced solution both passes the Lean checker and satisfies the algorithmic instruction.
\framework{} attains the best solve rate on every (benchmark, model) cell, with the absolute gain over the stronger baseline ($\Delta$) ranging from +4.6 to +11.2 percentage points.
The baselines split by benchmark type: \textsc{Plain} is competitive on textbook-style Verina and AlgoVeri, whereas \textsc{Seq} consistently beats \textsc{Plain} on repository-grounded Lean4Commit0 by 1.8--3.7 points.
The positive $\Delta$ across all three regimes (solve rates 59--77\% on Verina, 31--48\% on AlgoVeri, 9--22\% on Lean4Commit0) shows that joint planning improves over the two primary baselines regardless of benchmark difficulty.

\begin{table}[t]
  \centering
  \caption{Implementation-planning ablation with Claude-Opus-4.7 (solve rate, \%).}
  \label{tab:plan-seq}
  \setlength{\tabcolsep}{7pt}
  \footnotesize
  \begin{tabular}{lrrrrr}
    \toprule
    Benchmark & \textsc{Plain} & \textsc{Seq} & \textsc{Plan-Seq} & \framework{} & $\Delta$ \\
    \midrule
    Verina       & 68.8 & 68.3 & 69.8 & \textbf{74.6} & +4.8 \\
    AlgoVeri     & 39.0 & 40.3 & 44.8 & \textbf{48.1} & +3.3 \\
    Lean4Commit0 & 13.9 & 17.6 & 13.9 & \textbf{22.2} & +8.3 \\
    \bottomrule
  \end{tabular}
  \vspace{-8pt}
\end{table}

Table~\ref{tab:plan-seq} isolates joint planning from generic implementation planning: \framework{} outperforms \textsc{Plan-Seq} by 3.3--8.3 points.
\textsc{Plan-Seq} helps on Verina and AlgoVeri but hurts on Lean4Commit0, where independently reasonable API implementations may be difficult to connect under a relational theorem unless they are planned against their shared proof obligations.

\textbf{Plan viability.}
Across all 189 Verina traces for \framework{} with Claude-Opus-4.7, 131 of 141 successes (92.9\%) retain the initial plan, while only 4 of 48 failures (8.3\%) are attributed to an inadequate plan; Appendix~\ref{app:plan-analysis} reports the full breakdown and caveat that retention is only a proxy for proof-burden prediction accuracy.

To compare efficiency on tasks that require non-trivial search, we restrict to the top 25\% of tasks by mean cost across the three methods (n=48 on Verina, n=20 on AlgoVeri, n=27 on Lean4Commit0).
Appendix Table~\ref{tab:top25-cost-time} reports per-task cost and wall-clock time on this difficult subset.
\framework{} is the cheapest and fastest method on every (benchmark, model) cell, reducing cost by 3.0--39.6\% and wall-clock time by 3.1--37.2\% relative to the better baseline for each metric.
The largest savings appear on Verina and AlgoVeri, while Lean4Commit0 leaves less room for time reduction because all three methods often run close to the per-task limit.
On the all-solved intersection, \framework{} remains broadly comparable to the baselines, with some visible planning overhead on easier cells (Appendix~\ref{app:cost-time-intersect}).

Trace inspection attributes much of \textsc{Seq}'s cost to \emph{full-restart repair}: after local repairs fail under the committed program, the agent rewrites both artifacts. \framework{} checks structural consistency before elaboration and restarts less often; Appendix~\ref{app:full-restart} gives the breakdown.

\subsection{Case Study}
\label{sec:case:llrbt}

On AlgoVeri's left-leaning red-black-tree deletion task, \textsc{Seq}'s structural route reaches 6{,}344 lines without closing, while rebuild closes in 1{,}176 lines only after inheriting redundant invariants. \framework{} compares both sketches, recognizes that rebuild reduces the value-set goal to list permutation, and closes it in 1{,}105 lines and about 34 minutes.

A repository-derived counterpart is \textsc{Memchr}, a three-API relational task solved only by \framework{} under the same backend and budget (Appendix~\ref{app:memchr}).

\textbf{Scope and limitations.}
Evidence is restricted to Lean~4, closed-source backends, Python-skewed core APIs, imperfect specification gates, and single runs; transfer to other verifiers, open models, and whole systems remains open (Appendix~\ref{app:limitations}).

\section{Related Work}
\label{sec:related}

{\bf Verified code generation.}
Proof-carrying code~\citep{necula1997proof} cast safe execution as a producer-consumer protocol with a trusted checker; LLM-driven verified code generation instead makes the LLM the producer, shifting the difficulty from certificate \emph{checking} to certificate \emph{construction}.
Across SMT and ITP systems~\citep{lattuada2023verus,leino2010dafny,swamy2016dependent,moura2021lean,bertot2004coq,nipkow2002isabelle}, recent agents construct or repair certificates after the implementation is fixed~\citep{yang2025autoverus,yang2025verusage,yang2026exverus,yan2025reform,first2023baldur,thakur2024copra,gloeckle2026wybecoder}.
Proof-search systems add decomposition before local proving~\citep{li2026goedel,zhang2026planning} or scale neuro-symbolic search to system-level developments~\citep{he2026neurosymbolic}, but still begin after the program and obligations are fixed; theorem-proving priors~\citep{azerbayev2024llemma,xin2024deepseekprover,ren2025deepseekproverv2,hubert2025alphaproof} leave that schedule unchanged.
The Floyd--Hoare and refinement-synthesis traditions~\citep{floyd1967assigning,hoare1969axiomatic,dijkstra1976discipline,gries1981science,polikarpova2016synquid} long argued instead that program structure and correctness argument should be developed together; our workflow ports this discipline into LLM-driven verified code generation, with the plan playing the role of the inductive assertion or refinement schema before either the program or the certificate is elaborated.
Explicit planning has also been shown to help LLM agents on \emph{unverified} code generation~\citep{jiang2024self,wang2023plan}; the new requirement in our setting is that the plan align the program decomposition with the proof decomposition the verifier will demand.

{\bf Benchmark.}
Existing benchmarks are either short, problem-level verified-generation tasks~\citep{ye2025verina,thakur2025clever,zhao2026algoveri,vericoding2025} or annotation reconstruction over fixed programs~\citep{loughridge2024dafnybench,yang2025autoverus,yang2025verusage}.
Lean4Commit0 differs on three axes from prior verified-code-generation benchmarks (see Appendix~\ref{app:bench-comparison} for a side-by-side comparison): each task is a library-level bundle of core APIs derived from a real GitHub project rather than a single function; specifications include relational properties across multiple APIs of the same library; and admission is gated by a uniform automated pipeline (reference satisfaction, mutation rejection, LLM review), so reported solve rates can be read against a known floor on spec strength.

\section{Conclusion}
\label{sec:conclusion}

Committing to implementation structure before proof structure induces brittle repair loops; \framework{} instead fixes a shared decomposition, bridging predicate, and auxiliary lemmas before elaboration, routing feedback to local repair or explicit replanning.
Lean4Commit0 complements this workflow with repository-derived, library-level tasks, relational API specifications, and a uniform quality pipeline, moving verified generation toward correctness arguments planned with the code rather than recovered after the fact.

\bibliographystyle{unsrtnat}
\bibliography{references}

\newpage
\appendix


\section{LLM Usage Disclosure}
\label{app:llm-disclosure}

Large language models are the central object of study in this work; their use as experimental backends (Codex-GPT-5.5, Gemini-3-Pro, Claude-Sonnet-4.6, and Claude-Opus-4.7) is fully documented in Section~\ref{sec:experiments}. Beyond this experimental role, the authors used LLM-based assistants (ChatGPT and Claude) for limited, non-substantive support during manuscript preparation, specifically polishing English phrasing in already-drafted paragraphs, suggesting LaTeX formatting fixes, and aiding in routine code refactoring for the evaluation harness. All research ideas, the design of \framework{}, the construction of Lean4Commit0, the experimental protocol, the analysis of results, and all claims made in this paper were conceived, verified, and written by the authors, who take full responsibility for the final content.

\section{Broader Impacts}
\label{app:broader-impacts}

We discuss the societal implications of \framework{} and Lean4Commit0 along three axes.

\textbf{Intended positive impact.}
The work targets \emph{verified} code generation, where every accepted output ships with a machine-checked correctness proof against a formal specification. To the extent that planning-first agent workflows raise the solve rate of such pipelines, the deployable consequence is a higher fraction of LLM-produced software whose correctness is established by Lean's kernel rather than by sampled tests, which is particularly relevant for safety-critical and security-critical settings (cryptographic primitives, parsers, configuration logic, smart-contract code). Lean4Commit0 itself contributes a denser supply of repository-derived, library-level training and evaluation tasks for the formal-methods community, with relational specifications across APIs that are largely absent from prior single-function benchmarks.

\textbf{Risks and misuse pathways.}
We do not foresee direct misuse pathways beyond the residual risks already shared with general code-generation systems, namely the acceleration of software development in directions that may themselves be undesirable (proprietary lock-in, capability concentration, automation of work whose social distribution remains unsettled). A more specific risk is \emph{misplaced trust}: a verified artifact is correct only with respect to the shipped specification, so an under-constrained or vacuous specification can yield code that the verifier accepts but that does not satisfy the user's actual intent. Lean4Commit0 mitigates this for benchmark specifications via the three-stage quality pipeline of \S\ref{sec:benchmark:quality} (reference satisfaction, mutation rejection, LLM review), but downstream users who write their own specifications retain full responsibility for adequacy.

\textbf{Data and privacy.}
The benchmark surfaces only Lean specifications, reference implementations, and tests derived from public open-source libraries already redistributed via Commit0~\citep{zhao2024commit0}, with their original licenses preserved; it introduces no scraping of private content, no human-subject data, and no surveillance dimension. The released agent harness operates on this same public corpus and on user-supplied Lean files, and does not collect or transmit user data beyond what the underlying LLM API providers already see.

\section{Limitations}
\label{app:limitations}

We discuss the main limitations of this work along three axes.

\textbf{Substrate scope.}
\framework{} and Lean4Commit0 are instantiated exclusively on Lean~4. Other verified-programming substrates, in particular SMT-backed systems such as Dafny and Verus and dependently typed languages such as F$^\star$ and Rocq, differ substantially in proof-obligation granularity, automation profile, and the cost of failed attempts. In Verus, a shared plan would pair Rust control and data structures with pre/postconditions, loop invariants, decreases clauses, ghost state, and SMT-facing auxiliary lemmas; Dafny admits an analogous mapping. Local annotation or solver errors could be repaired under the fixed plan, whereas a missing shared invariant or incompatible decomposition would trigger replanning. This transfer is plausible but not empirically established here.

\textbf{Benchmark and specification coverage.}
Lean4Commit0 is constructed from real GitHub libraries surfaced through Commit0, but the extraction pipeline focuses on \emph{core APIs} rather than full project compilation, and the resulting tasks therefore underrepresent very long-horizon dependencies, complex build configurations, and dynamic features that resist Lean modeling. The selected libraries also reflect Commit0's domain distribution and are skewed toward Python-origin code, which may not generalize to systems-heavy or numerically intensive software. In addition, Lean4Commit0 specifications are constructed by our pipeline rather than fully hand-written. Each candidate is filtered by reference satisfaction, behavior-changing mutation rejection, and LLM review, but a specification could still be subtly under-constrained relative to a use case the reference tests do not exercise. Solve rate is therefore a function of the specifications we ship, not of the underlying intent.

\textbf{Statistical, cost, and backend considerations.}
Because each $(\text{task},\text{model},\text{method})$ configuration is run once, we do not estimate within-task stochastic variance or perform formal significance testing across all settings. The absolute costs reported are tied to the API pricing in effect at evaluation time and should be read as relative comparisons rather than fixed budgets. Our experiments also use four frontier closed-source LLM backends (Codex-GPT-5.5, Gemini-3-Pro, Claude-Sonnet-4.6, Claude-Opus-4.7); it remains open whether the planning advantage transfers to open-weight models, smaller models, or specialized fine-tuned theorem-proving models.

\section{Full \texorpdfstring{\lean{listMax}}{listMax} Example}
\label{app:listmax}

This appendix gives the complete Lean~4 source for the \lean{listMax} example of \S\ref{sec:problem}.
Both versions are checked against the same specification \lean{Spec}; the file compiles under Lean~4 with Mathlib.

\textbf{Shared specification.}
The postcondition states that the returned value lies in the input list and upper-bounds every element.

\vspace{0.35em}
\begin{leancode}
def Spec (xs : List Int) (m : Int) : Prop :=
  m ∈ xs ∧ ∀ x ∈ xs, x ≤ m
\end{leancode}
\vspace{0.35em}

\textbf{Version (A): structural recursion.}
The induction mirrors the spec: \lean{Spec} itself serves as the bridging predicate, and each conjunct of the postcondition is discharged case by case.

\vspace{0.35em}
\begin{leancode-green}
def listMaxA : (xs : List Int) → xs ≠ [] → Int
  | [a],          _ => a
  | a :: b :: rs, _ => max a (listMaxA (b :: rs) (by simp))

theorem listMaxA_spec :
    ∀ (xs : List Int) (h : xs ≠ []), Spec xs (listMaxA xs h)
  | [a], _ => by
      refine ⟨by simp [listMaxA], ?_⟩
      intro x hx; simp at hx; subst hx; simp [listMaxA]
  | a :: b :: rs, _ => by
      have ih := listMaxA_spec (b :: rs) (by simp)
      obtain ⟨ih_mem, ih_ub⟩ := ih
      refine ⟨?_, ?_⟩
      · show listMaxA (a :: b :: rs) _ ∈ a :: b :: rs
        simp only [listMaxA]
        rcases le_total a (listMaxA (b :: rs) (by simp)) with hle | hle
        · rw [max_eq_right hle]; simp [ih_mem]
        · rw [max_eq_left hle]; simp
      · intro x hx
        simp only [listMaxA]
        rcases List.mem_cons.mp hx with rfl | hx'
        · exact le_max_left _ _
        · exact le_max_of_le_right (ih_ub x hx')
\end{leancode-green}
\vspace{0.35em}

\textbf{Version (B): accumulator fold.}
The fold cannot satisfy \lean{Spec} directly.
The proof must first generalize over an arbitrary accumulator \lean{init}, discharge the auxiliary lemma \lean{foldl_max_gen}, and then specialize \lean{init} to \lean{xs.head!}.
The lemma is a fact about \lean{foldl}, not about the maximum function.

\vspace{0.35em}
\begin{leancode-blue}
def listMaxB (xs : List Int) (h : xs ≠ []) : Int :=
  xs.foldl max xs.head!

lemma foldl_max_gen :
    ∀ (init : Int) (ys : List Int), Spec (init :: ys) (ys.foldl max init)
  | init, [] => by
      refine ⟨by simp, ?_⟩
      intro z hz; simp at hz; subst hz; simp
  | init, y :: t => by
      have ih := foldl_max_gen (max init y) t
      obtain ⟨ih_mem, ih_ub⟩ := ih
      refine ⟨?_, ?_⟩
      · show (y :: t).foldl max init ∈ init :: y :: t
        simp only [List.foldl]
        rcases List.mem_cons.mp ih_mem with hcase | hcase
        · rw [hcase]
          rcases le_total init y with hle | hle
          · rw [max_eq_right hle]; simp
          · rw [max_eq_left hle]; simp
        · simp [hcase]
      · intro z hz
        simp only [List.foldl]
        rcases List.mem_cons.mp hz with heq | hz'
        · rw [heq]
          exact (le_max_left init y).trans
            (ih_ub _ (List.mem_cons.mpr (Or.inl rfl)))
        · rcases List.mem_cons.mp hz' with heq | hz''
          · rw [heq]
            exact (le_max_right init y).trans
              (ih_ub _ (List.mem_cons.mpr (Or.inl rfl)))
          · exact ih_ub z (List.mem_cons.mpr (Or.inr hz''))

theorem listMaxB_spec (xs : List Int) (h : xs ≠ []) :
    Spec xs (listMaxB xs h) := by
  match xs, h with
  | a :: rs, _ =>
      have hg := foldl_max_gen a rs
      simpa [listMaxB, List.head!, Spec] using hg
\end{leancode-blue}
\vspace{0.35em}

\section{Prompt Templates}
\label{app:prompts}

This appendix reproduces the methodology-defining portions of the agent skill that drives \framework{}'s planning and elaboration phases. Planning and elaboration enter the same skill at different phases of the workflow shown below: planning halts at Phase~3 (the plan gate) and emits the plan file plus a sketch of code and proof obligations; elaboration takes over from Phase~4 to fill the implementation and proof bodies. For brevity we omit operational scaffolding (LSP-tool listings, subagent routing, fallback scripts, troubleshooting, and the cross-reference index between sub-skills); these portions are not load-bearing for the algorithmic claims of the paper. The anonymized agent harness, prompts, benchmark, and evaluation scripts are available in the Figshare snapshot at \url{https://figshare.com/s/2ab7020cacaffcf0d258}.

\textbf{Skill description.}
\begin{Verbatim}[fontsize=\scriptsize,frame=single,framesep=2pt,xleftmargin=0pt,xrightmargin=0pt,samepage=false]
Lean 4 code generation & verification with a co-derivation discipline -- code and
proof sketch are derived together from a single shared reasoning trace. Use when
editing .lean files with sorry placeholders.
\end{Verbatim}

\textbf{Core principles.}
\begin{Verbatim}[fontsize=\scriptsize,frame=single,framesep=2pt,xleftmargin=0pt,xrightmargin=0pt,samepage=false]
- Never modify specifications. Preconditions, postconditions, and theorem
  statements are read-only.

- Respect Instructions. A `/-! Instruction: -/` docstring or `instruction.md` is
  a hard constraint -- do not take spec-satisfying shortcuts that violate the
  stated algorithm or complexity.

- Plan commitment before writing code & proof. Code emitted in Phase 4 must
  realise the plan's `## Algorithmic idea` and `## Library support`. If you find
  yourself typing a different algorithm, that is plan failure -- revert and
  re-enter Phase 2. No silent algorithm switch at implementation time -- the
  single most common path to the spec-copy / wrong-complexity cheat.

- Standard axioms only. `propext`, `Classical.choice`, `Quot.sound`,
  `Lean.ofReduceBool`, `Lean.trustCompiler`. Nothing else. In particular, do not
  use `Classical.choose` on an existence proof to extract a value the
  Instruction says must be computed -- that is not an algorithm.

- Search during the plan, not during the proof. Phase 2 commits the plan to
  specific stdlib functions and mathlib lemmas (`## Library support`); Phase 4
  code and Phase 5 proof both implement those same commitments. If Phase 5 needs
  a lemma the plan didn't commit to, treat it as plan failure -> horizontal
  pivot back to Phase 2.
\end{Verbatim}

\textbf{Workflow.}
The skill enforces six phases, each producing an artifact and gating on an explicit exit check; do not advance until the check holds.
\begin{Verbatim}[fontsize=\scriptsize,frame=single,framesep=2pt,xleftmargin=0pt,xrightmargin=0pt,samepage=false]
1. Understand
   - Produces: classified sorry inventory (code vs proof); spec + instruction
     restated.
   - Exit: every sorry enumerated; instruction source located or confirmed
     absent.

2. Co-derive plan
   - Produces: <file>.plan.md with the schema's required sections.
   - Exit: plan file exists on disk; all required sections present; any
     locked-algorithm instruction is restated verbatim under
     `## Restated contract`. May not edit the .lean file before this artifact.

3. Plan gate
   - Produces: external PASS-PLAN / FAIL verdict from `lean_plan_verify` (or
     the `lean-verifier-plan` subagent).
   - Exit: verdict is PASS-PLAN. On FAIL, revise the plan file and re-run the
     same gate. Self-review is not a gate.

4. Emit code sketch + proof sketch
   - Produces: code bodies written in full; proof bodies left as `sorry`.
     `decreasing_by sorry` may stay (or be omitted if Lean auto-derives).
     Plus stub statements for the main theorem and any auxiliary lemmas.
   - Exit: `lean_diagnostic_messages` reports no errors; expected `sorry`
     warnings only; signatures elaborate so the sketch parses.

5. Fill tactic details
   - Produces: all tactic-level sorries closed (proof sorries and any
     `decreasing_by sorry` -- both are tactic blocks).
   - Exit: no errors and no in-scope sorries.

6. Quality gate (final)
   - Produces: go / no-go verdict on delivery.
   - Exit: `lean-verifier-mechanical` returns PASS-MECHANICAL and
     `lean-verifier-instruction` returns PASS-INSTRUCTION; act on each
     verdict's Action: line otherwise.
\end{Verbatim}

\textbf{Plan schema (Phase~2 artifact).}
The plan file at \texttt{<file>.plan.md} must contain the following sections; each is the explicit commitment that Phase~4 code and Phase~5 proof are required to realise.
\begin{Verbatim}[fontsize=\scriptsize,frame=single,framesep=2pt,xleftmargin=0pt,xrightmargin=0pt,samepage=false]
## Restated contract
  Verbatim restatement of the formal spec and any algorithmic / asymptotic
  instruction.

## Algorithmic idea
  The decomposition: recursion variable, case split, termination measure.

## Library support
  Specific stdlib functions and mathlib lemma names the plan commits to.
  Adding a lemma here that does not exist is a Phase-2 failure, not a Phase-5
  failure.

## Proof obligation sketch
  For the main theorem and each auxiliary lemma, the bridging predicate or
  invariant under which Phase 5 will close it.
\end{Verbatim}

\textbf{Phase 6 -- Quality gate.}
Two delegated checks run in sequence (stop-on-first-failure); both must pass before delivery.
\begin{Verbatim}[fontsize=\scriptsize,frame=single,framesep=2pt,xleftmargin=0pt,xrightmargin=0pt,samepage=false]
1. Mechanical gate: lean-verifier-mechanical.
   Runs sorry scan, compile check, axiom check.
   On FAIL, parent re-enters Phase 5 targeting the failing sorry / error.

2. Instruction compliance: lean-verifier-instruction.
   Reads the @start code ... @end code region and judges whether the
   implementation satisfies the Instruction's complexity / algorithmic intent.
   On FAIL, parent reverts and re-enters Phase 2 with a different algorithm
   or stdlib pivot, then redoes Phase 3 -> 4 -> 5 -> 6.

Do not swap the implementation in place and re-run Phase 6 -- that bypasses
the co-derivation gate and desyncs plan, code, and proof.
\end{Verbatim}

\textbf{When stuck.}
Two recovery axes -- vertical (refine the current plan, once or twice max) and horizontal (discard the plan, re-enter Phase~2). Patching a doomed plan is the more common failure mode this skill creates than rewriting one; when in doubt, pivot.
\begin{Verbatim}[fontsize=\scriptsize,frame=single,framesep=2pt,xleftmargin=0pt,xrightmargin=0pt,samepage=false]
Hard caps per target (one sorry or one compile error), per Phase-5 cycle:
  - Max 6 edits OR 8 distinct tactic candidates, whichever first.
  - Max 2 attempts per (file, line, error_or_sorry_id) signature with no
    error_count decrease.

Pivot horizontally as soon as any of these fire:
  - Same signature in 2 consecutive iterations, no error_count decrease.
  - Same sub-goal failed 2 distinct tactics AND 1 invariant rewording.
  - Phase 2: 2 consecutive empty searches for the same commitment -- the
    mathlib path the plan named doesn't exist; revise ## Algorithmic idea.
  - Phase 4-5: needing a lemma not in ## Library support -- the plan didn't
    anticipate the case; re-derive ## Library support and ## Algorithmic
    idea together, not mid-proof.
  - decreasing_by keeps failing -- recursion scheme has no clean termination
    measure.
  - Type-system fight (Fin vs Nat, getElem vs getElem!) eats > 5 turns --
    data-structure choice is wrong.
  - Postcondition is a multi-conjunction (>=3 independent claims) but the
    plan has only one main lemma.

Vertical first: refine tactic -> invariant. Allowed once or twice; if still
stuck, go horizontal: discard plan + Phase-4 sketch. Re-enter Phase 2 with
a different recursion variable, fuel formulation, split lemma, or different
data structure. Not wasted work -- it is the right move when the plan can't
carry the proof.
\end{Verbatim}

\section{Benchmark Comparison}
\label{app:bench-comparison}

Table~\ref{tab:bench-comparison} situates Lean4Commit0 alongside prior verified-code-generation benchmarks. Each benchmark is characterized along three axes:

\begin{itemize}[leftmargin=*, topsep=2pt, itemsep=2pt]
\item \textbf{Substrate.} The verifier the benchmark targets. ITP-based substrates (Lean~4) require explicit proof terms or tactic scripts, whereas SMT-based substrates (Dafny, Verus) discharge annotation bundles through Z3~\citep{de2008z3} or CVC5~\citep{barbosa2022cvc5}. ``Multi'' indicates that the benchmark provides tasks across more than one substrate.
\item \textbf{Task.} Which artifacts the model is asked to produce: \emph{spec + code + proof} synthesizes all three from a high-level natural-language description; \emph{code + proof} fixes the formal specification and asks the model to write the implementation together with the proof artifact; \emph{proof only} fixes both the implementation and the specification and asks the model to fill in the verifier annotations. The proof artifact subsumes both ITP tactic/term proofs and SMT annotation bundles.
\item \textbf{Source.} Where the underlying problems come from: manually curated short functions, problems adapted from HumanEval-style or competitive-programming sources, or real-world software corpora.
\end{itemize}

The existing benchmarks in this table all draw tasks from manually curated single-function problems or annotated SMT corpora. Lean4Commit0 instead draws from real GitHub libraries via Commit0~\citep{zhao2024commit0}, forming each task as a bundle of selected core APIs and admitting relational specifications across those APIs (\S\ref{sec:benchmark}); these structural differences are the reason a \emph{plan} that fixes the program decomposition before either the implementation or the proof is elaborated has visible leverage on this benchmark, as documented in Table~\ref{tab:main-solve}.

\begin{table}[t]
  \centering
  \setlength{\tabcolsep}{4pt}
  \caption{Verified code generation benchmarks. \emph{Task} indicates the artifact the model is asked to produce: \emph{spec + code + proof} (synthesize all three from a high-level description), \emph{code + proof} (formal specification is given; model writes the implementation and proof artifact), or \emph{proof only} (implementation and specification are given; model writes only the proof artifact). The proof artifact subsumes both ITP tactic/term proofs and SMT annotation bundles. Existing benchmarks draw tasks from manually curated single-function problems or annotated SMT corpora; Lean4Commit0 instead derives library-level core-API bundles from real GitHub libraries via Commit0~\citep{zhao2024commit0} and adds relational specifications across APIs.}
  \label{tab:bench-comparison}
  \small
  \begin{tabular}{l l l l}
    \toprule
    Benchmark & Substrate & Task & Source \\
    \midrule
    Verina~\citep{ye2025verina}                 & Lean~4 & spec + code + proof & manually curated tasks                  \\
    Clever~\citep{thakur2025clever}             & Lean~4 & spec + code + proof & adapted from HumanEval                  \\
    AlgoVeri~\citep{zhao2026algoveri}           & Multi  & code + proof        & classical algorithms (curated)          \\
    DafnyBench~\citep{loughridge2024dafnybench} & Dafny  & proof only          & GitHub Dafny + Clover + MBPP            \\
    AutoVerus~\citep{yang2025autoverus}         & Verus  & proof only          & translated from MBPP, Diffy, CloverBench \\
    VeruSAGE~\citep{yang2025verusage}           & Verus  & proof only          & open-source Verus-verified Rust systems \\
    Vericoding~\citep{vericoding2025}           & Multi  & code + proof        & 12,504 specs (HumanEval, APPS, etc.) \\
    \midrule
    \textbf{Lean4Commit0 (ours)}                & Lean~4 & code + proof        & real GitHub libraries         \\
    \bottomrule
  \end{tabular}
\end{table}

\section{Fabric Relational Specification Example}
\label{app:fabric}

This appendix shows the Lean~4 task excerpt used in \S\ref{sec:benchmark} to illustrate cross-API relational specifications.
Fabric is a Python library for SSH-based deployment and remote execution; in Lean4Commit0 we model core configuration behavior from its API.
The agent must fill every \lean{sorry}, including both implementation bodies and proofs.

\vspace{0.35em}
\begin{leancode-green}
def Config.set (cfg : Config) (key : String) (value : ConfigValue)
    (level : ConfigLevel) : Config := sorry

def Config.get (cfg : Config) (key : String) : Option ConfigValue := sorry
\end{leancode-green}
\vspace{0.35em}

\begin{leancode-blue}
@[reducible]
def Config.set_get_roundtrip_rel (cfg : Config) (k : String) (v : ConfigValue)
    (level : ConfigLevel) (result : Option ConfigValue) : Prop :=
  result = some v

@[reducible]
def Config.higher_level_overrides_rel (cfg : Config) (k : String)
    (v1 v2 : ConfigValue) (lo hi : ConfigLevel) : Prop :=
  lo.precedence < hi.precedence →
    ((cfg.set k v1 lo).set k v2 hi).get k = some v2
\end{leancode-blue}
\vspace{0.35em}

\begin{leancode-red}
theorem Config.set_get_roundtrip_spec (cfg : Config) (k : String)
    (v : ConfigValue) (level : ConfigLevel) :
    Config.set_get_roundtrip_rel cfg k v level ((cfg.set k v level).get k) := by
  sorry

theorem Config.higher_level_overrides_spec (cfg : Config) (k : String)
    (v1 v2 : ConfigValue) (lo hi : ConfigLevel)
    (hlt : lo.precedence < hi.precedence) :
    Config.higher_level_overrides_rel cfg k v1 v2 lo hi := by
  sorry
\end{leancode-red}
\vspace{0.35em}

\section{Lean4Commit0 \texorpdfstring{\textsc{Memchr}}{Memchr} Case Study}
\label{app:memchr}

The \textsc{Memchr} task contains three APIs---\lean{memchr}, \lean{memmem}, and \lean{memrchr}---and four correctness theorems, including a relational requirement that forward and reverse search agree on whether the target byte occurs.
Under Claude-Opus-4.7 and the same budget as the main experiment, neither primary baseline solves the task.
\textsc{Plain} cannot connect the available \lean{List.idxOf} lemmas to the optional-index specification before timing out, while \textsc{Seq} commits to an unbounded search and fails the final algorithmic evaluation.

\framework{} instead makes the finite witness domain explicit in the shared plan and aligns the first- and last-occurrence proofs with \lean{Nat.find} and \lean{Nat.findGreatest}.
This makes the cross-API theorem compositional: after extracting occurrence and bound facts from the forward-search specification, the reverse-search specification rules out the contradictory branch.
The central proof connection is:

\begin{leancode-green}
have hi := (memchr_spec needle haystack).1 i hmemchr
exact ((memrchr_spec needle haystack).2
  hmemrchr i hi.1 hi.2.1).elim
\end{leancode-green}

The converse direction is symmetric.
Lean accepts all three implementations and all four proofs without \lean{sorry} or custom axioms; the complete traces are included in the released artifact.

\section{Plan Retention and Failure Analysis}
\label{app:plan-analysis}

We inspect all 189 Verina traces for \framework{} with Claude-Opus-4.7.
Among 141 successful runs, 131 (92.9\%) retain the initially selected plan through completion and 10 (7.1\%) replace it after the original route proves infeasible.
Because a task may admit multiple valid plans, retention is an observable proxy for plan viability rather than a direct measure of perfect proof-burden prediction.

Among the 48 failed runs, 4 (8.3\%) fail because of an inadequate plan, 3 (6.3\%) during program elaboration, 15 (31.3\%) during proof elaboration under an otherwise viable plan, and 26 (54.2\%) because of timeout or tool failure.
Thus only 4 of 48 failures are attributed to plan inadequacy; most arise from proof elaboration or execution limits after a viable plan has been chosen.

\section{Efficiency on the Difficult Subset}
\label{app:difficult-efficiency}

Table~\ref{tab:top25-cost-time} reports the cost and wall-clock comparison summarized in Section~\ref{sec:experiments:main} for the top 25\% of tasks by mean cost.
\begin{table}[t]
  \centering
  \caption{Average cost (USD) and wall-clock time (minutes) per task on the \emph{difficult} subset of each benchmark (the top 25\% of tasks ranked by mean cost across \textsc{Plain}, \textsc{Seq}, and \framework{}). Lower is better.}
  \label{tab:top25-cost-time}
  \setlength{\tabcolsep}{4pt}
  \small
  \begin{tabular}{ll *{3}{>{\raggedleft\arraybackslash}p{1.2cm} >{\raggedleft\arraybackslash}p{1.0cm}}}
    \toprule
    \multirow{2.5}{*}{Benchmark} & \multirow{2.5}{*}{Model}
      & \multicolumn{2}{c}{\textsc{Plain}}
      & \multicolumn{2}{c}{\textsc{Seq}}
      & \multicolumn{2}{c}{\framework{}} \\
    \cmidrule(lr){3-4}\cmidrule(lr){5-6}\cmidrule(lr){7-8}
     & & Cost & Time & Cost & Time & Cost & Time \\
    \midrule
    \multirow{4}{*}{Verina}
      & Codex-GPT-5.5       & 12.7 &  65.5 &  13.3 &  63.8 & \textbf{11.3} &  \textbf{47.6} \\
      & Gemini-3-Pro      & 14.9 &  83.0 &  15.7 &  89.6 & \textbf{13.6} &  \textbf{52.1} \\
      & Claude-Sonnet-4.6 & 19.7 & 132.0 & 19.9 & 149.0 & \textbf{11.9} & \textbf{103.0} \\
      & Claude-Opus-4.7   & 25.0 & 164.5 & 26.6 & 162.2 & \textbf{23.0} & \textbf{140.5} \\
    \midrule
    \multirow{4}{*}{AlgoVeri}
      & Codex-GPT-5.5       & 11.3 &  91.2 & 13.8 &  72.8 & \textbf{9.1} &  \textbf{48.9}\\
      & Gemini-3-Pro      & 13.4 & 110.1 & 13.6 &  79.6 &  \textbf{8.4} &  \textbf{63.2} \\
       & Claude-Sonnet-4.6 & 29.4 & 134.3 & 26.2 & 137.1 & \textbf{24.2} & \textbf{125.9}  \\
      & Claude-Opus-4.7   & 26.1 & 158.0 & 21.3 & 116.6 & \textbf{15.6} &  \textbf{95.1} \\
    \midrule
    \multirow{4}{*}{Lean4Commit0}
      & Codex-GPT-5.5       & 16.5 &  90.6 & 18.1 & 113.4 & \textbf{12.6} &  \textbf{82.7} \\
      & Gemini-3-Pro      & 20.1 & 149.7 & 25.2 & 142.6  & \textbf{16.7} & \textbf{135.7} \\
      & Claude-Sonnet-4.6  & 20.0 & 149.6 & 28.2 & 160.5 & \textbf{18.2} & \textbf{141.6} \\
      & Claude-Opus-4.7   & 29.8 & 125.8 & 29.6 & 130.5 & \textbf{28.7} & \textbf{121.9} \\
    \bottomrule
  \end{tabular}
  \vspace{-10pt}
\end{table}

\section{Full-Restart Repair Frequency}
\label{app:full-restart}

Section~\ref{sec:experiments:main} described a \emph{full-restart repair} failure mode: after local repairs fail under the committed program structure, the agent abandons the current derivation and rewrites both code and proof from scratch, sometimes more than once. We measure the frequency of this failure mode by inspecting traces and counting runs that perform such a full restart; lower is better.

Table~\ref{tab:full-restart} breaks down this full-restart rate by benchmark and backend. \framework{} attains the lowest rate on every Verina and AlgoVeri cell, typically reducing the restart rate by a factor of two to four relative to the stronger baseline. On Lean4Commit0, all three methods are below 5\% because library-level relational specifications already pin down enough behavior to leave little room for full restarts.

\begin{table}[t]
  \centering
  \caption{Full-restart repair rate (\% of runs where local repair fails and the agent rewrites the code-and-proof derivation from scratch). Lower is better.}
  \label{tab:full-restart}
  \setlength{\tabcolsep}{6pt}
  \small
  \begin{tabular}{ll *{3}{>{\raggedleft\arraybackslash}p{1.2cm}}}
    \toprule
    Benchmark & Model & \textsc{Plain} & \textsc{Seq} & \framework{} \\
    \midrule
    \multirow{4}{*}{Verina ($n{=}189$)}
      & Codex-GPT-5.5     & 25.9 & 28.6 & \textbf{11.7} \\
      & Gemini-3-Pro      & 24.2 & 24.7 & \textbf{7.8}  \\
      & Claude-Sonnet-4.6 & 17.4 & 18.5 & \textbf{5.3}  \\
      & Claude-Opus-4.7   & 28.5 & 24.9 & \textbf{8.5}  \\
    \midrule
    \multirow{4}{*}{AlgoVeri ($n{=}77$)}
      & Codex-GPT-5.5     & 33.4 & 34.9 & \textbf{11.2} \\
      & Gemini-3-Pro      & 21.3 & 32.2 & \textbf{7.0}  \\
      & Claude-Sonnet-4.6 & 21.9 & 23.2 & \textbf{5.8}  \\
      & Claude-Opus-4.7   & 24.4 & 33.6 & \textbf{9.1}  \\
    \midrule
    \multirow{4}{*}{Lean4Commit0 ($n{=}108$)}
      & Codex-GPT-5.5     &  3.7 &  4.3 & \textbf{1.7}  \\
      & Gemini-3-Pro      &  4.1 &  4.9 & \textbf{1.6}  \\
      & Claude-Sonnet-4.6 &  3.2 &  4.4 & \textbf{1.2}  \\
      & Claude-Opus-4.7   &  4.3 &  5.2 & \textbf{2.2}  \\
    \bottomrule
  \end{tabular}
\end{table}

\section{Per-Task Efficiency on the Solved Intersection}
\label{app:cost-time-intersect}

Table~\ref{tab:cost-time-intersect} reports per-task cost and time on the intersection of tasks where Plain, Seq, and \framework{} all produce algorithmically-correct solutions, isolating per-task efficiency from coverage differences in the main results.

\begin{table}[t]
  \centering
  \caption{Average per-task cost (USD) and time (minutes) on the intersection of
  correctly-solved tasks (Plain, Seq, and \framework{} all yield algorithmically-correct
  solutions). Isolates per-task efficiency from coverage.}
  \label{tab:cost-time-intersect}
  \setlength{\tabcolsep}{4pt}
  \begin{tabular}{ll *{3}{>{\raggedleft\arraybackslash}p{1.2cm} >{\raggedleft\arraybackslash}p{1.0cm}}}
    \toprule
    \multirow{2.5}{*}{Benchmark} & \multirow{2.5}{*}{Model}
      & \multicolumn{2}{c}{\textsc{Plain}} & \multicolumn{2}{c}{\textsc{Seq}} & \multicolumn{2}{c}{\framework{}} \\
    \cmidrule(lr){3-4}\cmidrule(lr){5-6}\cmidrule(lr){7-8}
     & & Cost & Time & Cost & Time & Cost & Time \\
    \midrule
    \multirow{4}{*}{Verina}
      & Codex-GPT-5.5     & 1.96 & 9.5  & 1.61 & 8.6  & 1.71 & 9.2  \\
      & Gemini-3-Pro      & 2.24 & 13.5 & 1.71 & 9.3  & 1.62 & 10.7 \\
      & Claude-Sonnet-4.6 & 2.22 & 21.6 & 2.22 & 21.7 & 2.13 & 20.9 \\
      & Claude-Opus-4.7   & 2.55 & 18.1 & 2.63 & 19.1 & 2.67 & 19.6 \\
    \midrule
    \multirow{4}{*}{AlgoVeri}
      & Codex-GPT-5.5     & 2.78 & 13.9 & 2.95 & 16.8 & 1.20 & 19.6 \\
      & Gemini-3-Pro      & 2.67 & 20.5 & 4.17 & 18.1 & 1.10 & 19.2 \\
      & Claude-Sonnet-4.6 & 4.02 & 16.5 & 5.71 & 28.0 & 5.31 & 31.8 \\
      & Claude-Opus-4.7   & 2.31 & 27.8 & 2.31 & 26.6 & 2.86 & 36.9 \\
    \midrule
    \multirow{4}{*}{Lean4Commit0}
      & Codex-GPT-5.5     & 1.46 & 14.2 & 2.83 & 26.0 & 2.90 & 18.4 \\
      & Gemini-3-Pro      & 2.28 & 17.0 & 3.71 & 35.3 & 2.71 & 21.4 \\
      & Claude-Sonnet-4.6 & 4.01 & 32.5 & 3.89 & 41.2 & 3.95 & 30.5 \\
      & Claude-Opus-4.7   & 3.75 & 32.9 & 3.79 & 47.7 & 4.06 & 42.7 \\
    \bottomrule
  \end{tabular}
\end{table}

\end{document}